*Article*

# Scale-aware Neural Network for Semantic Segmentation of Multi-resolution Remote Sensing Images

Libo Wang [1], Ce Zhang [2,3], Rui Li [1], Chenxi Duan [4], Xiaoliang Meng [1,*] and Peter M. Atkinson [2]

1. School of Remote Sensing and Information Engineering, Wuhan University, Wuhan 430079, China; wanglibo@whu.edu.cn (L.W.); lironui@whu.edu.cn (R.L.); xmeng@whu.edu.cn (X.M.)
2. Lancaster Environment Centre, Lancaster University, Lancaster LA1 4YQ, United Kingdom; c.zhang9@lancaster.ac.uk (C.Z.); pma@lancaster.ac.uk (P.M.A.)
3. UK Centre for Ecology & Hydrology, Library Avenue, Lancaster LA1 4AP, United Kingdom; c.zhang9@lancaster.ac.uk (C.Z.)
4. Faculty of Geo-Information Science and Earth Observation (ITC), University of Twente, Enschede, the Netherlands; c.duan@utwente.nl
* Correspondence: xmeng@whu.edu.cn (X.M.)

**Abstract:** Assigning geospatial objects with specific categories at the pixel level is a fundamental task in remote sensing image analysis. Along with rapid development in sensor technologies, remotely sensed images can be captured at multiple spatial resolutions (MSR) with information content manifested at different scales. Extracting information from these MSR images represents huge opportunities for enhanced feature representation and characterisation. However, MSR images suffer from two critical issues: 1) increased scale variation of geo-objects and 2) loss of detailed information at coarse spatial resolutions. To bridge these gaps, in this paper, we propose a novel scale-aware neural network (SaNet) for semantic segmentation of MSR remotely sensed imagery. SaNet deploys a densely connected feature network (DCFFM) module to capture high-quality multi-scale context, such that the scale variation is handled properly and the quality of segmentation is increased for both large and small objects. A spatial feature recalibration (SFRM) module is further incorporated into the network to learn intact semantic content with enhanced spatial relationships, where the negative effects of information loss are removed. The combination of DCFFM and SFRM allows SaNet to learn scale-aware feature representation, which outperforms the existing multi-scale feature representation. Extensive experiments on three semantic segmentation datasets demonstrated the effectiveness of the proposed SaNet in cross-resolution segmentation.

**Keywords:** deep convolutional neural network; multiple spatial resolutions; remote sensing; scale-aware feature representation; semantic segmentation.

## 1. Introduction

Fine spatial resolution (FSR) remotely sensed images are characterised by rich spatial information and detailed objects with semantic content. Semantic segmentation using FSR remotely sensed imagery has been a hot topic in the remote sensing community, which essentially undertakes a dense pixel-level classification task and has been applied in various geo-related applications including land cover classification [1], infrastructure planning [2], and territorial management [3], as well as change detection [4] and other urban applications [5] [6,7].

Driven by rapid development in sensor technology over the past few years, FSR remotely sensed images are captured increasingly at multiple spatial resolutions (MSR), meaning that FSR remotely sensed images are shifting towards MSR remotely sensed images [8]. MSR remotely sensed images provide much richer detailed information and more various geometrical characterisation than FSR images [9,10]. Meanwhile, diverse spatial resolutions bring complex scale variation of geospatial objects as illustrated in Fig 1. Thus, semantic segmentation of MSR remotely sensed images is an extremely challenging task but with profound impacts.

To handle the multi-scale variation in MSR semantic segmentation, existing research relies on two major strategies: (1) methods based on the traditional handcrafted features and (2) methods based on hierarchical feature representations of deep convolutional neural network (DCNN) [11]. Traditional hand-crafted methods involve either two-stage segmentation or one-stage segmentation. The multi-resolution segmentation algorithm is the most successful two-stage segmentation approach [12], which partitions an image into homogeneous segments in the first stage and assigns these segments into particular categories during the second stage [13,14]. To capture the scale variation of geo-objects, MRS-based methods introduce a manually controlled scale parameter for determining the object size [15,16]. One-stage handcrafted approaches consider segmentation as a patch-based dense classification task in computer vision. Typically a handcrafted feature extractor such as the scale-invariant feature transformer (SIFT) is adopted to extract multi-scale patterns within MSR images [17,18]. These well-engineered features are fed into supervised classifiers such as support vector machine (SVM) [19], random forests [20] [21], and conditional random fields (CRF) [22] to realise pixel-level semantic labelling or segmentation. However, designing effective hand-crafted features is time-consuming and the performance of handcrafted features depends on parameter settings and specific data, thus limiting its generalisation capability.

Deep convolutional neural networks have brought significant breakthroughs in semantic segmentation [23-26], thanks to their hierarchical feature representation in an end-to-end and automatic fashion [27]. The learned hierarchical features are highly robust and generalised, by which the multi-scale variation can be captured and characterized [28]. Common DCNN based semantic segmentation of multi-scale objects includes image pyramid, multi-level feature fusion (MFF) framework, and the spatial pyramid pooling (SPP) architecture.

The image pyramid method trains parallel networks with input images at several resolutions and merges multi-resolution features together [29]. Although it could strengthen multi-scale feature representation, the complex training process involves high computational complexity, reducing the efficiency of the network. To enhance the multi-scale representation of deep networks without increasing extensive computational complexity [30], the MFF framework and the SPP architecture have been investigated frequently in recent years [31-33]. The MFF framework merges low-level detailed features and high-level semantic features by skip connections to establish multi-scale representation. For example, U-Net and its variants concatenate encoding features and decoding features via skip connections, and the merged features are able to restore to the original image resolution [34-36]. Feature pyramid network (FPN) series build an extra top-down pathway to integrate multi-scale features [37,38]. The SPP architecture develops parallel atrous convolution layers or pooling layers for field-of-view enlargement, thereby avoiding the disadvantage of the fixed receptive field in the traditional convolutional layer and capturing multi-scale information. Typical models include PSPNet [39], Deeplab [40-42] and its extensions in the remote sensing domain [43,44].

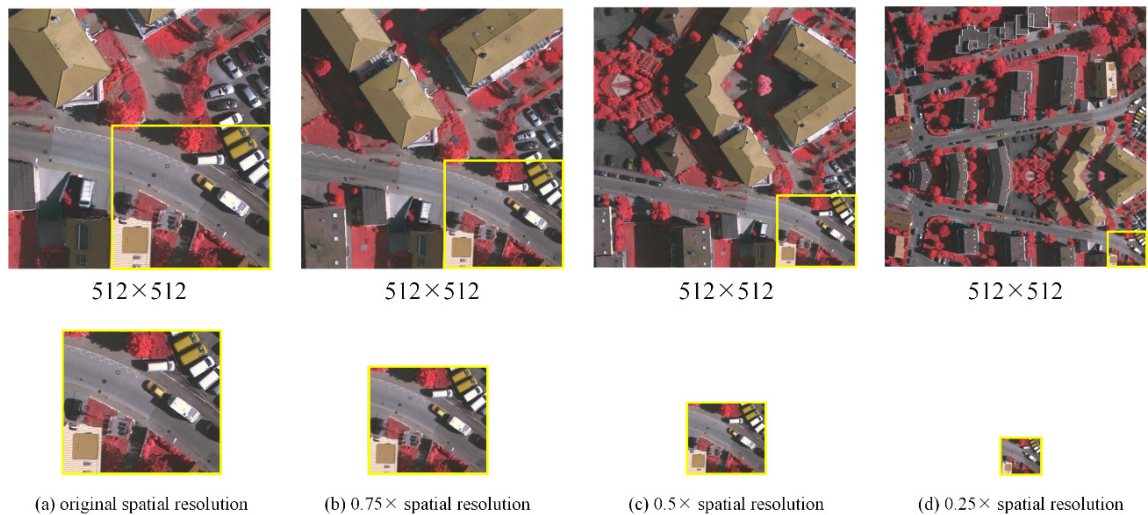

**Figure 1.** Illustration of the complex scale variation of geospatial objects in MSR remotely sensed images. Each image is a 512×512 px patch. Due to the diverse spatial resolutions, the scale variation within (e.g. vehicles in the highlighted yellow region) and between objects (e.g. buildings and vehicles) has been enlarged.

Although these methods have achieved significant advancement in semantic segmentation of multi-scale objects, they still demonstrate limited quality and fidelity for segmenting MSR remotely sensed images. The main reasons are two-fold: (1) Those approaches with a single multi-scale representation are difficult to model the complex scale variation of geo-objects in MSR remotely sensed images. (2) the methods are less effective in abstracting geo-objects due to ignoring the loss of details in objects at coarse spatial resolutions. As shown in Figure 1, the details of vehicles (e.g. window) are clear at the original spatial resolution, but much unclear at the 0.25× spatial resolution.

In this paper, we propose a novel scale-aware neural network (SaNet) for semantic segmentation of MSR remotely sensed images. Specifically, we explore the multi-scale structure and propose a novel densely connected feature fusion module (DCFFM). To avoid the limitation of the single multi-scale representation, the DCFFM module combines the advantages of the MFF framework and SPP architecture for high-quality multi-scale representation. It constructs several dense connections with different enlarged receptive field sizes to capture rich multi-scale information in the fashion of SPP. Most importantly, weighted fusion (WF) operations are employed for multi-level feature fusion, correcting the latent fitting residual from semantic gaps in features at different levels. Moreover, we present a spatial feature recalibration module (SFRM) that models the scale-invariant spatial relationship within semantic features of geo-objects to strengthen the feature extraction at coarse resolutions. The SFRM builds a dual-branched structure to model spatial relationships at different scales, which is particularly suitable for multi-resolution images. With the combination of the DCFFM and SFRM, SaNet could extract the scale-aware feature to capture the complex scale variation for semantic segmentation of MSR remotely sensed images. The structure of the proposed SaNet is elegantly designed and separable, which can be easily transplanted into other DCNNs trained end-to-end automatically. The major contributions of this paper are summarized as follows:

(1) A novel scale-aware neural network is proposed for semantic segmentation of MSR remotely sensed images. It learns scale-aware feature representation instead of current multi-scale feature representation to address the large scale variation of geo-objects in MSR remotely sensed images.

(2) We develop a simple yet effective spatial feature recalibration module with a dual-branched structure. It enhances the scale-invariant feature representation by modelling the spatial relationship within semantic features, providing a new perspective for alleviating the effects of loss in object details at coarse resolutions.

(3) We propose a densely connected feature fusion module to obtain high-quality multi-scale representation. To leverage the advantage of the SPP architecture in multi-scale information capture, we design the large field connection to enlarge the receptive field of high-level features for further connecting with features at different levels. In addition, we employ weighted fusion operations for multi-level feature aggregation. It increases the generalization of fused features significantly by reducing the latent fitting residual.

The remainder of this paper is organized as follows. The architecture of SaNet and its components are detailed in Section 2. Experimental comparison between SaNet and a set of benchmark methods are provided in Section 3. A comprehensive discussion is presented in Section 4. Finally, conclusions are drawn in Section 5.

## 2. Materials and Methods

### 2.1. Overview

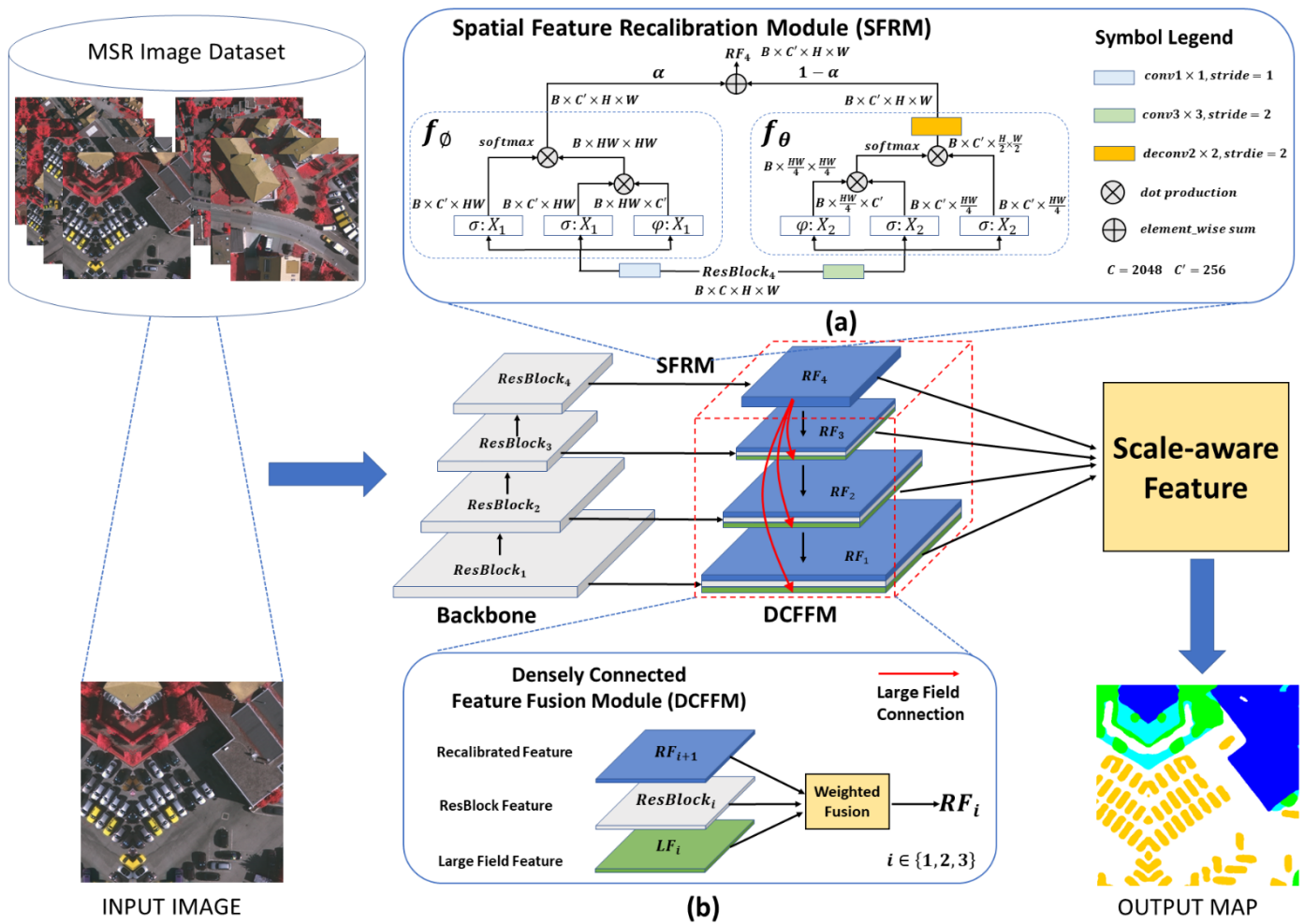

**Figure 2.** The architecture of the proposed SaNet.

The overall architecture of the proposed SaNet is composed of the ResNet backbone [45], SFRM and DCFFM, as shown in **Figure 2**. The ResNet backbone consists of four residual blocks, extracting the corresponding four ResBlock features from the input image: $ResBlock_1$, $ResBlock_2$, $ResBlock_3$ and $ResBlock_4$. Especially, $ResBlock_4$ is downscaled 16 times and its size is the same as $ResBlock_3$. Considering the efficiency of SaNet, SFRM is only deployed on top of the ResNet backbone to recalibrate the high-level semantic feature $ResBlock_4$. Thereafter, the recalibrated feature $RF_4$ is fed into DCFFM. DCFFM employs three large field connections (marked by red arrows) to generate the large field features (i.e., $LF_1$, $LF_2$ and $LF_3$) from $RF_4$, and constructs a top-down pathway to aggregate ResBlock features (grey), recalibrated features (blue) and large field features (green) by weighted fusion operations. The three layers of DCFFM produce three recalibrated features (i.e., $RF_1$, $RF_2$ and $RF_3$) which are then fused with $RF_4$ to generate the scale-aware feature for final segmentation.

### 2.2. Spatial relationship enhancement with SFRM

To address the issue of object detail loss at coarse spatial resolutions, we design a spatial feature recalibration module that could model invariant spatial relationships within semantic features of geospatial objects, thereby increasing the feature representation for MSR images. SFRM constructs two branches of different sizes to model the global spatial relationship at diverse scales, remedying the limitation of the single branch structure that can only extract the fixed-scale spatial interactions. As shown in **Figure 2 (a)**, the input is the high-level feature $ResBlock_4$ and the output is the recalibrated feature $RF_4$. By applying two convolution layers with different kernel sizes and strides, the input $ResBlock_4$ is divided into two branch features, $X_1$ and $X_2$. The generation process of $X_1$ and $X_2$ can be formalized as follows:

$$X_1 = ResBlock_4 \cdot W_1[C, C', K_1, S_1] \tag{1}$$

$$X_2 = ResBlock_4 \cdot W_2[C, C', K_2, S_2] \tag{2}$$

where $K_1 = 1$ and $S_1 = 1$ represent the size and stride of the filter $W_1$. Similarly, $K_2 = 3$ and $S_2 = 2$ represent the size and stride of the filter $W_2$. $C = 2048$ and $C' = 256$ denote the input channels and output channels, respectively. Operated by the two convolution layers, the size of $X_1$ is twice the size of $X_2$.

The generated $X_1$ and $X_2$ are then fed into the spatial functions $f_\emptyset$ and $f_\theta$, respectively, to provide information on the global spatial relationship. Thereafter, the spatially enhanced features are merged by a weighted element-wise sum operation to generate $RF_4$:

$$RF_4(X_1, X_2) = \alpha \cdot f_\emptyset(X_1) + (1 - \alpha) \cdot f_\theta(X_2) \tag{3}$$

Here, $\alpha$ is a trainable variable to suppress redundant features produced during the merging process. The spatial functions $f_\emptyset$ and $f_\theta$ can be defined as:

$$f_\emptyset(X_1) = \emptyset \left( \sigma(X_1) \cdot f_s(\varphi(X_1) \cdot \sigma(X_1)) \right) \tag{4}$$

$$f_\theta(X_2) = \theta \left( \sigma(X_2) \cdot f_s(\varphi(X_2) \cdot \sigma(X_2)) \right) \tag{5}$$

where $f_s$ represents the softmax activation function. The detailed implementation steps of $f_\emptyset$ and $f_\theta$ are as follows:

1) The feature map $X_1$ is reshaped by $\sigma$ and $\varphi$ into $C' \times HW$ and $HW \times C'$, respectively. Similarly, the feature map $X_2$ is reshaped into $C' \times \frac{HW}{4}$ and $\frac{HW}{4} \times C'$.
2) A dot production operation is applied to $\sigma(X_1)$ and $\varphi(X_1)$ to produce the spatial relationship matrix $HW \times HW$, which is further fed into the softmax activation function $f_s$ to generate the probability map for feature recalibration. Meanwhile, $\sigma(X_2)$ and $\varphi(X_2)$ are processed by a similar procedure, but the shape of the corresponding probability map is $\frac{HW}{4} \times \frac{HW}{4}$.
3) The probability map is multiplied by $\sigma(X_1)$ to generate the spatial recalibrated feature of $X_1$. The spatial recalibrated feature of $X_2$ is generated in the same way. Further, the operation $\emptyset$ resizes the spatial recalibrated feature of $X_1$ to $C' \times H \times W$, while the combined operation $\theta$ deploys a deconvolution layer to upsample the spatial recalibrated feature of $X_2$ and then resize it to $C' \times H \times W$.

Processed by the two branches, the recalibrated feature $RF_4$ is abundant in information relating to the global spatial relationships and capable of capturing intact semantic content from coarse-resolution images.

### 2.3. High-quality multi-scale representation with DCFFM

We proposed a novel feature aggregation module for abstracting multi-scale geo-objects from MSR remotely sensed images. Since this novel module utilizes the large field connections to densely connect the multi-level features, we name it densely connected feature fusion module. The main advantage of DCFFM is to capture high-quality multi-scale contexts through a weighted fusion of semantic features at different sizes and receptive fields. The structure of DCFFM is illustrated in the dashed red box of **Figure 2**.

*Large feild connection*: To match the structure of the ResNet backbone, we designed three large field connections in DCFFM. Each connection contains a 2-D atrous convolution to generate the corresponding large field feature ($LF_i$) from the recalibrated feature $RF_4$, whereafter the stacked transposed convolutions are adopted to control the output size when necessary. The large field connection is defined as a function with the following equation:

$$LF_i(RF_4) = T_{3-i} \circ D_i(RF_4), \qquad i \in \{1, 2, 3\} \tag{6}$$

where $i$ denotes the layer index. $T^\circ$ is a resize function that performs a 2×2 transpose convolution with stride 2 repeatedly, and $3-i$ denotes the number of repetitions. $D_i$ denotes a 2-D atrous convolution for receptive field enlargement, which can be defined as:

$$D_i(RF_4) = \sum_{k_1=1}^{K} \sum_{k_2=1}^{K} RF_4[m + f_r(i) \cdot k_1, n + f_r(i) \cdot k_2] \cdot w[k_1, k_2] \tag{7}$$

where $[m, n]$ and $[k_1, k_2]$ represent the spatial position indices of the output $D_i$ and the convolution filter $w$, respectively. Here, $K$ is set to 3. $f_r(i)$ denotes the dilated rate of $D_i$ which can be computed as follow:

$$f_r(i) = 24 - 6i \tag{8}$$

According to the different layer $i$, three large field connections could generate three large field features with diverse receptive fields and sizes, providing richer multi-scale contexts for feature fusion.

*Weighted fusion*: The three generated large field features are aggregated with the corresponding ResBlock features and recalibrated features by a weighted element-wise sum operation to strengthen the generalization capability of fused features, as exhibited in **Figure 2 (b)**. The equation is as follows:

$$RF_i = \begin{cases} RF_i & \text{if } i = 4 \\ \alpha_1 \cdot f_\mu(RF_{i+1}) + \alpha_2 \cdot f_\delta(ResBlock_i) + \alpha_3 \cdot LF_i, & \text{if } i \in \{1,2,3\} \end{cases} \tag{9}$$

where $f_\mu$ is a resize operation to unify the shape of $RF_{i+1}$ and $LF_i$, while $f_\delta$ is a standard 1×1 convolution to unify the channels of $ResBlock_i$ and $LF_i$. $\alpha_1, \alpha_2, \alpha_3$ denote the weight coefficients and always satisfy $\alpha_1 + \alpha_2 + \alpha_3 = 1$.

Eventually, to capitalise on the benefits provided by spatial relationship enhancement and high-quality multi-scale representation, we further merge $RF_1$, $RF_2$, $RF_3$, $RF_4$ to generate the scale-aware feature ($SF$) for final segmentation. The formula is as follows:

$$SF = RF_1 + RF_2 + RF_3 + RF_4 \tag{10}$$

## 3. Results

*3.1. Experimenal settings*

3.1.1. Implementation Details

All models in the experiments are implemented with PyTorch framework on a single NVIDIA GTX 2080ti GPU with a batch size of 4. For fast convergence, we deploy the AdamW optimizer to train all models in the experiments. The base learning rate is set to 1e-4 and the weight decay value was 0.01. The early stopping technique is applied to control the training time for preventing overfitting. Cross-entropy loss is chosen as the loss function. Please note that only scale-invariant image transformation (random flip) is used for data augmentation to avoid the influence of the scale variations.

3.1.2. Models for comparison

To test the cross-resolution generalization capability of the proposed SaNet, we select various competitive methods for comparison, including multi-scale feature aggregation models like the feature pyramid network (FPN) [37] and pyramid scene network (PSPNet) [39], the multi-view context aggregation method Deeplabv3+ [41], and the criss-cross attention network (CCNet) [46], as well specially designed models for semantic labelling of remotely sensed images, such as relational context-aware fully convolutional network (S-RA-FCN) [47], the dense dilated convolutions merging network (DDCM-Net) [43], edge-aware neural network (EaNet) [44], MACUNet [48] and MAResUNet [49]. Besides, ablation studies are conducted with the following model design:

(1) Baseline: An upsampling operation is employed on top of the backbone to construct the single-scale network Baseline. The feature maps produced by the Baseline are restored directly to the same size as the original input image.

(2) Baseline+SRM and Baseline+SFRM: The spatial relationship module (SRM) [47] and our SFRM are added into the Baseline to construct two spatial relationship networks (i.e., Baseline+SRM and Baseline+SFRM).

(3) Baseline+FPN and Baseline+DCFFM: The FPN module [37] and our DCFFM are embedded into the Baseline to construct two multi-scale networks (i.e., Baseline+FPN and Baseline+DCFFM).

3.1.3. Evaluation metrics

The performance of the proposed method is evaluated by the F1 score and overall accuracy, which can be calculated based on an accumulated confusion matrix:

$$precision = \frac{1}{k}\sum_{c=1}^{k}\frac{TP_c}{TP_c + FP_c} \quad (11)$$

$$recall = \frac{1}{k}\sum_{c=1}^{k}\frac{TP_c}{TP_c + FN_c} \quad (12)$$

$$F1 = 2 \times \frac{precison \times recall}{precision + recall} \quad (13)$$

$$OA = \frac{\sum_{c=1}^{k}TP_c}{N} \quad (14)$$

$$mIoU = \frac{1}{k}\sum_{c=1}^{k}\frac{TP_c}{TP_c + FP_c + FN_c} \quad (15)$$

where *c* represents the index of the class, $k$ denotes the number of classes. $TP_c$, $TN_c$, $FP_c$ and $FN_c$ indicate samples of true positives, true negatives, false positives, and false negatives of class *c*, respectively. $N$ is the total number of pixels in all classes.

*3.2. Experiments I: results on the LandCover.ai dataset*

LandCover.ai is a large-scale multi-resolution aerial imagery dataset for semantic segmentation [50], which collected true orthophoto RGB image tiles of 216.27 km² rural areas across Poland (a medium-sized country in Central Europe) under various optical and seasonal conditions. There are 33 image tiles (ca. 9000×9500 px) with a spatial resolution of 0.25 m and 8 image tiles (ca. 4200×4700 px) with a spatial resolution of 0.5 m, covering 176.76 km² and 39.51 km² respectively. The image tiles were manually annotated into four classes of geospatial objects, including water, building, woodland, and background.

In our experiments, the 33 image tiles with a spatial resolution of 0.25 m are randomly split into a training set (70%) and a validation set (30%). The 8 image tiles with a spatial resolution of 0.5 m are chosen as the test set. All image tiles are cropped into 512×512 px patches.

3.2.1. Ablation study on the Landcover.ai dataset

To evaluate the performance of the SFRM and DCFFM modules separately in the semantic mapping of MSR aerial images, we choose ResNet101 as the backbone and conduct ablation experiments.

*Ablation study for the spatial feature recalibration module*: Since the diversity in spatial resolution, a certain gap exists between the validation set and the test set (**TABLE 1**). Notably, with the employment of SRM and SFRM, the average mIoU increases by 8.6% and 10.4%, and the gap of mIoU reduced by 6% and 8.6%, compared to the Baseline. The lower gap indicates the stronger adaptivity of the model to spatial resolution. These results not only suggest that modelling spatial relationships could enhance feature representation of coarse-resolution images, but also show the superiority of our SFRM.

*Ablation study for densely connected feature fusion module:* As illustrated in TABLE 1, the deployment of FPN and DCFFM produces higher average mIoU scores (78.5% and 82.1%) and smaller gaps (12.4% and 8.2%), compared to the Baseline (71.0% and 17.4%), demonstrating the effectiveness of multi-scale representation for semantic labelling of MSR images as well as the advantage of DCFFM.

**Table 1.** Ablation Study for the SFRM and DCFFM. The backbone is ResNet101. The spatial resolution of the validation set and test set are 0.25 and 0.5 m, respectively. The values in bold are the best.

| Method | mIoU | | Avg | Gap |
|---|---|---|---|---|
| | Val set (0.25m) | Test set (0.5m) | | |
| Baseline | 79.7 | 62.3 | 71.0 | 17.4 |
| Baseline+SRM | 85.3 | 73.9 | 79.6 | 11.4 |
| Baseline+SFRM | 85.8 | 77.0 | 81.4 | 8.8 |
| Baseline+FPN (FPN) | 84.7 | 72.3 | 78.5 | 12.4 |
| Baseline+DCFFM | 86.2 | 78.0 | 82.1 | 8.2 |
| Baseline+SFRM+DCFFM (SaNet) | **88.2** | **81.2** | **84.7** | **7.0** |

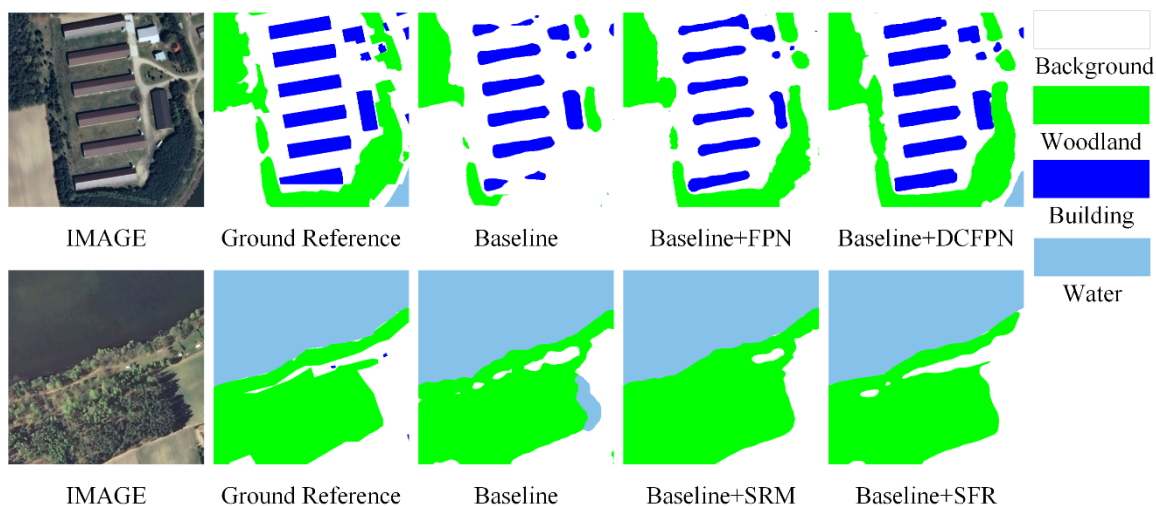

**Figure 3**. Visualization results of the ablation study of SFRM and DCFFM on the Landcover.ai test set.

The proposed SaNet maintains the highest average mIoU and the lowest gap, profiting from the simultaneous employment of SFRM and DCFFM. For a comprehensive comparison, the effectiveness and superiority of SFRM and DCFFM are visualized in **Figure 3**.

3.2.2. Comparison with state-of-the-art models

To further test the proposed SaNet for cross-resolution segmentation, numerous excellent benchmark methods are selected for comparison. Experimental results demonstrate that our SaNet maintains the highest average F1 score (93.5%) on the validation set (**TABLE 2**). Most importantly, the proposed SaNet still achieves the top mean F1score (89.1%) on the test set despite the spatial resolution coarsens (**TABLE 3**). The highest average F1 score and the lowest gap of F1 score also demonstrate the greater generalization capability of our SaNet in cross-resolution segmentation (**Figure 4**). Besides, the proposed SaNet is at least 5.5% higher than other methods in the F1-building score on the test set (**TABLE 3**). It segments buildings accurately, whereas other benchmark approaches depict coarse-structured and incomplete buildings (**Figure 5**)

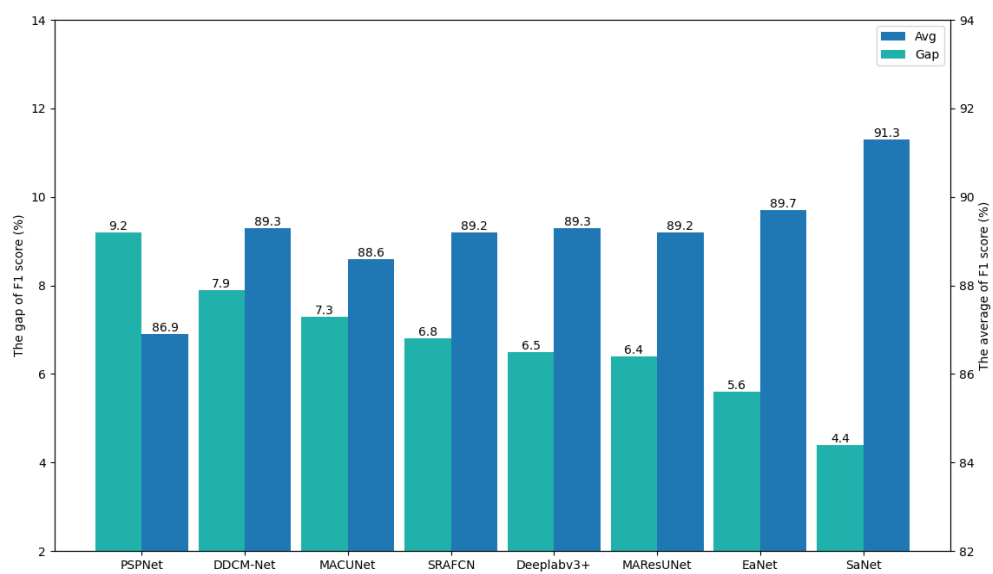

**Figure 4**. The average and gap of F1 score on the validation set and test set.

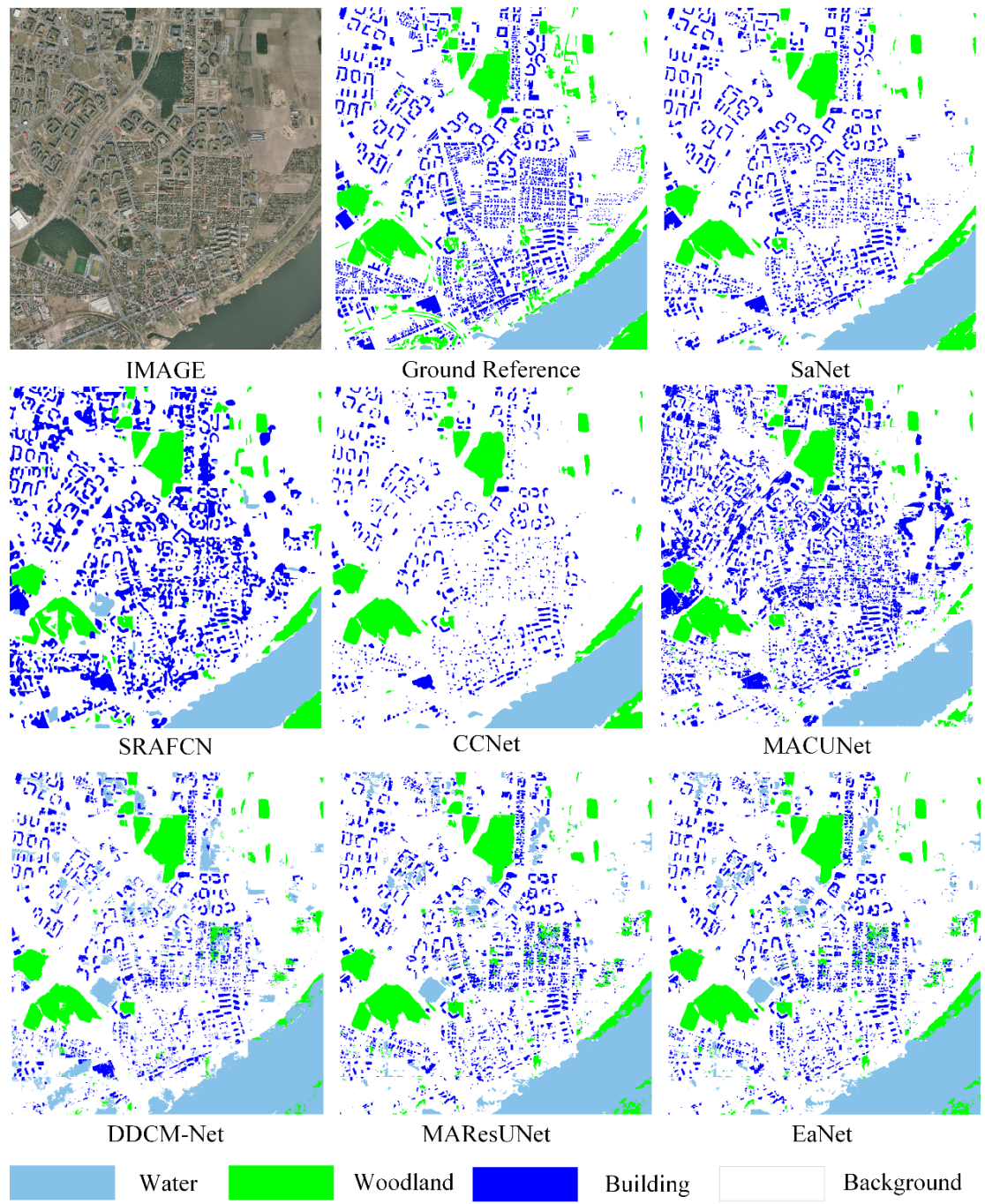

**Figure 5**. Visual comparison between our SaNet and other benchmark models on the Landcover.ai test set. The image size is 4232×4683.

**TABLE 2.** Quantitative comparison with state-of-the-art models on the validation set. The spatial resolution is 0.25m. The best values are in bold.

| Method | Backbone | F1-score | | | | Mean F1-score |
|---|---|---|---|---|---|---|
| | | Water | Building | Woodland | Background | |
| PSPNet | ResNet101 | 95.9 | 78.9 | 94.7 | 96.6 | 91.5 |
| Deeplabv3+ | ResNet101 | 96.3 | 82.8 | 94.4 | 96.5 | 92.5 |
| CCNet | ResNet101 | 95.9 | 80.8 | 94.7 | 96.5 | 92.0 |
| SRAFCN | VGG16 | 96.4 | 83.1 | 94.4 | 96.6 | 92.6 |
| DDCM-Net | ResNet101 | 96.9 | 84.4 | 94.6 | 96.6 | 93.2 |
| EaNet | ResNet101 | 96.1 | 82.5 | 94.7 | 96.6 | 92.5 |
| MACUNet | UNet | 95.7 | 82.2 | 94.3 | 96.5 | 92.2 |
| MAResUNet | UNet | 96.0 | 82.7 | 94.4 | 96.6 | 92.4 |
| SaNet | ResNet101 | 96.3 | **86.3** | **94.8** | **96.7** | **93.5** |

**TABLE 3.** Quantitative comparison with state-of-the-art models on the test set. The spatial resolution is 0.5m. The best values are in bold.

| Method | Backbone | F1-score | | | | Mean F1-score |
|---|---|---|---|---|---|---|
| | | Water | Building | Woodland | Background | |
| PSPNet | ResNet101 | 97.2 | 52.3 | 90.8 | 88.7 | 82.3 |
| Deeplabv3+ | ResNet101 | 96.8 | 67.8 | 90.9 | 88.6 | 86.0 |
| CCNet | ResNet101 | 97.2 | 58.2 | 91.4 | 89.1 | 84.0 |
| SRAFCN | VGG16 | 96.4 | 67.2 | 91.1 | 88.6 | 85.8 |
| DDCM-Net | ResNet101 | 97.1 | 64.1 | 91.0 | 88.9 | 85.3 |
| EaNet | ResNet101 | 96.9 | 68.9 | 92.1 | 89.8 | 86.9 |
| MACUNet | UNet | 96.5 | 67.1 | 89.2 | 86.9 | 84.9 |
| MAResUNet | UNet | 97.4 | 70.1 | 89.1 | 87.3 | 86.0 |
| SaNet | ResNet101 | 96.6 | **75.6** | **93.3** | **90.8** | **89.1** |

*3.3. Experiments II: results on the MSR Vaihingen dataset*

3.3.1. MSR Vaihingen dataset

The Vaihingen dataset consists of 33 very fine spatial resolution TOP image tiles at an average size of 2494×2064 pixels. Each TOP image tile has three multispectral bands (Near Infrared, Red, Green) as well as the digital surface model (DSM) and the normalized digital surface model (NDSM) with a 9 cm ground sampling distance (GSD). Only TOP image tiles were used in our experiments without DSM. The dataset involves five foreground classes (impervious surface, building, low vegetation, tree, car) and one background class (clutter). Following the recommendation of the previous work [43], 16 image tiles are selected as the training set and the remaining 17 image tiles as the original Vaihingen test set.

Particularly, to assess the performance of SaNet at a wider variety of spatial resolutions, we built the MSR Vaihingen dataset. Specifically, we first resample the image tiles of the original Vaihingen test set into 0.75 times manually and then crop them into 512×512 px patches, to generate the 0.75× Vaihingen test set. The 0.5× and 0.25× Vaihingen test sets are produced in the same fashion. The training set is cropped into 512×512 px patches directly. Data details are listed in **Table 4**.

**TABLE 4**. Details of the MSR Vaihingen dataset

| Dataset | Spatial Resolution (cm) | Patch Size (pixels) | Patch Numbers |
| --- | --- | --- | --- |
| Train set | 9 | 512×512 | 1092 |
| Original test set | 9 | 512×512 | 398 |
| 0.75× test set | 12 | 512×512 | 230 |
| 0.5× test set | 18 | 512×512 | 113 |
| 0.25× test set | 36 | 512×512 | 38 |

**TABLE 5.** Ablation Study for the SFRM module and DCFFM module. The backbone is ResNet101. Original, 0.75×, 0.5× and 0.25× represent the four Vaihingen test sets in Table 4.

| Method | OA | | | | Mean OA |
| --- | --- | --- | --- | --- | --- |
|  | original | 0.75× | 0.5× | 0.25× | |
| Baseline | 88.3 | 82.9 | 76.2 | 59.6 | 76.8 |
| Baseline+FPN (FPN) | 89.6 | 85.5 | 80.3 | 65.2 | 80.2 |
| Baseline+DCFFM | 89.8 | 86.0 | 81.0 | 66.2 | 80.8 |
| Baseline+SRM | 89.7 | 85.6 | 80.6 | 67.0 | 80.7 |
| Baseline+SFRM | 90.2 | 85.9 | 81.3 | 69.8 | 81.8 |
| Baseline+SFRM+DCFFM (SaNet) | 91.0 | 87.1 | 83.1 | 72.5 | 83.4 |

3.3.2. Ablation study on the MSR Vaihingen dataset

To evaluate the performance of the SFRM and DCFFM at more diverse spatial resolutions, we choose ResNet101 as the backbone and conduct ablation experiments on the MSR Vaihingen dataset.

*Ablation study for densely connected feature fusion module*: As listed in **TABLE 5**, compared to the Baseline, the utilization of FPN and DCFFM produces a significant increase in the mean OA (3.4% and 4.0%), which demonstrates the validity of multi-scale representation and the superiority of our DCFFM in comparison with FPN.

*Ablation study for the spatial feature recalibration module*: With the application of SRM and SFRM, the mean OA increased by 3.9% and 5.0%, respectively, compared to Baseline (**TABLE 5**). Particularly, Baseline+SFRM outperforms Baseline by 10.2% and exceeds Baseline+SRM by 2.8% on the 0.25×Vaihingen test set. These results suggest that enhancing information on the global spatial relationship could strengthen the adaptability of the network to MSR images. The significant increase in accuracy demonstrates the advantage of our SFRM in modelling spatial relationships. Moreover, by combining DCFFM and SFRM, our SaNet maintains the highest OA on the four Vaihingen test sets.

3.3.3. Comparison with other models

To further test the proposed SaNet for cross-resolution segmentation, we choose ResNet101 as the backbone network and compared SaNet with other excellent models on the four Vaihingen test sets. The Baseline+SRM is also selected for comparison as a competitive spatial relationship network. Experimental results demonstrate that the proposed SaNet outperforms other models in both mean F1 score (77.4%) and OA (83.4%) (**TABLE 6**). Specifically, SaNet increases the average OA by 6.6%, 2.7%, and 2.6% in comparison with Baseline, Baseline+SRM and Deeplabv3+, respectively. Meanwhile, SaNet produces increments of 2.4% and 2.0% in the average F1 score compared with DDCM-Net and EaNet.

With decreased spatial resolution, Baseline with a single-scale representation declines at the fastest rate, as shown in **Figure 6 (a)**, followed by the current multi-scale representation networks FPN, PSPNet, and Deeplabv3+ as well as the spatial relationship network Baseline+SRM and the specially designed networks DDCM-Net and EaNet. In contrast, the accuracy of SaNet reduces at the slowest rate. The OA secants (dashed lines) in **Figure 6 (b)** represent the declining magnitude of the OA when the spatial resolution decreases from the original to 0.25×. The proposed SaNet produces the smallest rate of decline.

As shown in **Figure 7**, the proposed SaNet achieves the most accurate segmentation maps compared with other methods. Particularly, the semantic content of the impervious surface is characterized effectively. Meanwhile, the complex contour of buildings is preserved completely on the 0.25× image.

**TABLE 6.** Quantitative comparison on the four Vaihingen test sets. The backbone is ResNet101. The values in bold are the best.

| Method | F1-score | | | | OA | | | | Mean F1-score | Mean OA |
|---|---|---|---|---|---|---|---|---|---|---|
| | original | 0.75× | 0.5× | 0.25× | original | 0.75× | 0.5× | 0.25× | | |
| Baseline | 84.9 | 76.4 | 65.9 | 48.2 | 88.3 | 82.9 | 76.2 | 59.6 | 68.9 | 76.8 |
| Baseline+SRM | 87.7 | 80.6 | 70.3 | 53.6 | 89.7 | 85.6 | 80.6 | 67.0 | 73.1 | 80.7 |
| FPN | 88.0 | 81.7 | 72.1 | 53.3 | 89.6 | 85.5 | 80.3 | 65.2 | 73.8 | 80.2 |
| PSPNet | 87.0 | 79.8 | 69.9 | 52.3 | 89.6 | 85.2 | 79.6 | 64.8 | 72.3 | 79.8 |
| Deeplabv3+ | 88.7 | 81.8 | 72.5 | 54.0 | 90.1 | 85.8 | 80.9 | 66.5 | 74.3 | 80.8 |
| DDCM-Net | 89.6 | 82.0 | 72.4 | 55.9 | 90.6 | 86.0 | 81.4 | 68.6 | 75.0 | 81.7 |
| EaNet | 89.8 | 82.6 | 73.4 | 55.9 | 90.7 | 86.1 | 81.2 | 68.0 | 75.4 | 81.5 |
| SaNet (ours) | **90.3** | **84.3** | **75.9** | **59.2** | **91.0** | **87.1** | **83.1** | **72.5** | **77.4** | **83.4** |

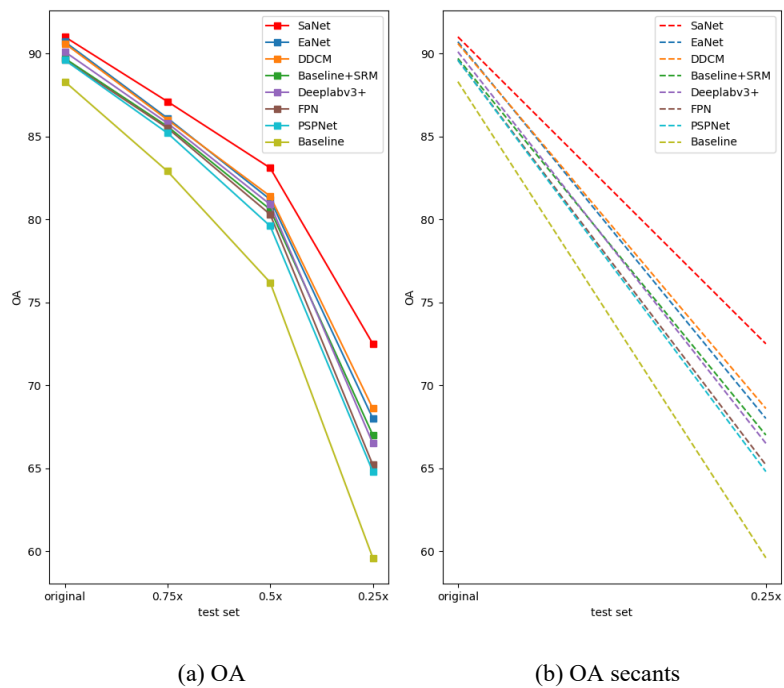

(a) OA    (b) OA secants

**Figure 6** (a) OA vs. the four Vaihingen test sets. (b) Secants of the OA, indicating the declining magnitude when spatial resolution decreases from the original to 0.25×.

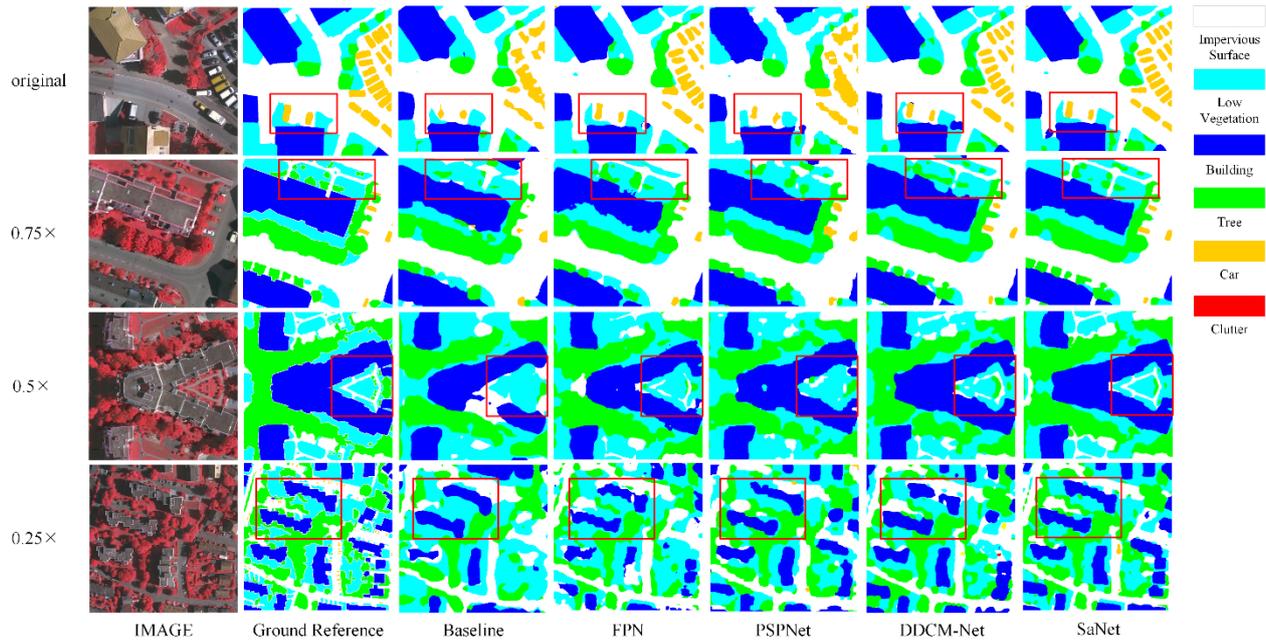

**Figure 7** Visual comparison of the four Vaihingen test sets. Each image is a 512×512 px patch. Control the zoom tool to >=200% to get a better view.

*3.4. Experiments III: results on the MSR Potsdam dataset*

3.4.1. MSR Potsdam dataset

The Potsdam dataset contains 38 very fine resolution TOP image tiles (GSD 5cm) at a size of 6000×6000 pixels and involves the same category information as the Vaihingen dataset. Four multispectral bands (Red, Green, Blue, and Near Infrared), as well as a DSM and NDSM, are provided in the dataset. The 24 image tiles are chosen as the training set, and the remaining tiles are selected as the original Potsdam test set. We utilize only TOP image tiles with three multispectral bands (Near Infrared, Red, Green) in the experiments. Notably, we create the MSR Potsdam dataset using the same strategy as the MSR Vaihingen dataset, as listed in **Table 7**.

**TABLE 7**. Details of the MSR Potsdam dataset

| Dataset | Spatial Resolution (cm) | Patch Size (pixels) | Patch Numbers |
|---|---|---|---|
| Train set | 5 | 512×512 | 3456 |
| Original test set | 5 | 512×512 | 2016 |
| 0.75× test set | 6.67 | 512×512 | 1134 |
| 0.5× test set | 10 | 512×512 | 504 |
| 0.25× test set | 20 | 512×512 | 126 |

3.4.2. Comparison with other models

The MSR Potsdam dataset is larger than the MSR Vaihingen dataset and more complex in terms of spatial details. We carry out comprehensive experiments on the four Potsdam test sets in **TABLE 7** to test the multi-resolution generalization capability of SaNet. Despite the intricate details in the images, our SaNet maintains superiority in both the mean F1 score (80.4%) and mean OA (83.4%) (**TABLE 8**). Particularly, SaNet exceeds all multi-scale contextual information aggregation methods, including Deeplabv3+, DDCM-Net, and EaNet, by 3.3%, 2.3%, and 2.2% in mean OA. For the 0.25× Potsdam test set, our SaNet delivers a respectable OA (69.7%) and F1 score (58.4%), outperforming the sub-optimal model EaNet by a large margin of 3.0% in the F1 score. The above-mentioned accuracy demonstrates the effectiveness and robustness of our SaNet for semantic labelling of MSR remotely sensed images. Moreover, SaNet yields the smallest declining magnitude (**Figure 8 (a)**) and achieves the most gentle rate of decline in accuracy with coarsening spatial resolution (**Figure 8 (b)**).

The segmentation results are shown in **Figure 9**, where regions with obvious improvement are marked by red boxes. The proposed SaNet with DCFFM and SFRM exhibits the smoothest visual appearance with the least red clutter noise, as shown in the first row of **Figure 9**. Labelling buildings with scale-aware features extracted by SaNet is more capable of recreating the complete object. For example, SaNet recognizes the complete, regular shape of the main building as shown in the second row of **Figure 9**, where other methods draw out the building into an incomplete and irregular semantic object due to the interference of impervious surface. In the 0.25× image, the SaNet represents the geometries of two adjacent buildings in the red box region effectively, whereas other approaches identify them as a single object (fourth row of **Figure 9**). Meanwhile, small objects like cars are also identified accurately in the third row of **Figure 9**.

**TABLE 8.** Quantitative comparison on the four Potsdam test sets. The backbone is ResNet101. The values in bold are the best.

| Method | F1-score | | | | OA | | | | Mean F1-score | Mean OA |
|---|---|---|---|---|---|---|---|---|---|---|
| | original | 0.75× | 0.5× | 0.25× | original | 0.75× | 0.5× | 0.25× | | |
| Baseline | 87.8 | 82.1 | 73.5 | 47.9 | 86.9 | 83.5 | 77.9 | 58.3 | 72.8 | 76.7 |
| Baseline+SRM | 90.4 | 85.4 | 76.8 | 55.0 | 89.2 | 86.4 | 81.5 | 65.6 | 76.9 | 80.7 |
| FPN | 90.4 | 85.9 | 78.0 | 52.1 | 88.9 | 86.2 | 81.4 | 63.9 | 76.6 | 80.1 |
| PSPNet | 90.5 | 85.2 | 76.1 | 52.8 | 89.5 | 86.3 | 80.8 | 62.0 | 76.2 | 79.7 |
| Deeplabv3+ | 90.0 | 85.4 | 77.8 | 51.3 | 88.8 | 86.1 | 81.3 | 64.0 | 76.1 | 80.1 |
| DDCM-Net | 91.7 | 87.3 | 76.4 | 55.0 | 90.1 | 87.2 | 82.4 | 64.5 | 77.6 | 81.1 |
| EaNet | 91.9 | 87.1 | 78.8 | 55.4 | 90.4 | 87.2 | 82.3 | 65.0 | 78.3 | 81.2 |
| SaNet (ours) | **92.3** | **88.3** | **82.4** | **58.4** | **90.9** | **88.4** | **84.7** | **69.7** | **80.4** | **83.4** |

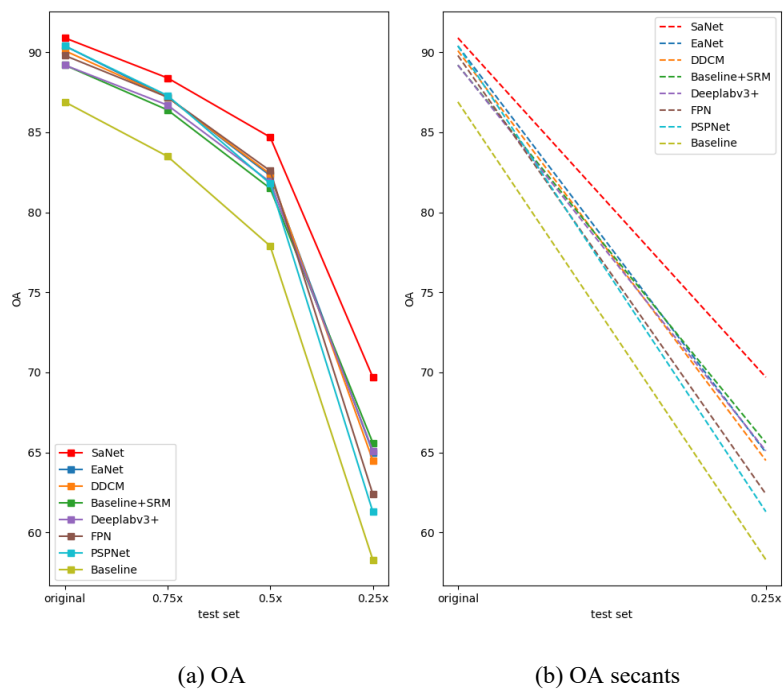

(a) OA          (b) OA secants

**Figure 8** (a) OA vs. the four Potsdam test sets. (b) Secants of the OA, denoting the declining magnitude when spatial resolution decreases from the original to 0.25×.

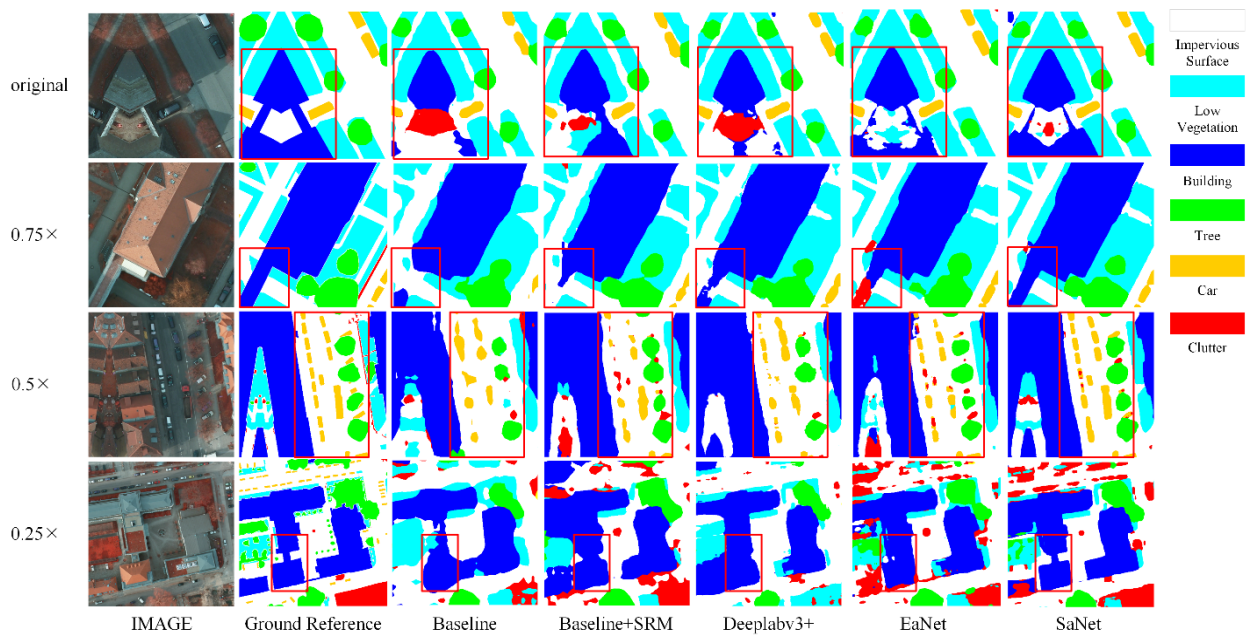

**Figure 9** Visual comparison of the four Potsdam test sets. Each image is a 512×512 px patch. Control the zoom tool to >=200% to get a better view.

## 4. Discussion

### 4.1. Influence of multiple spatial resolutions

Multiple spatial resolutions (MSR) in remotely sensed imagery bring several challenges to the existing semantic segmentation methods. Through extensive experiments, in this research, we discuss the influence of coarse spatial resolution in detail.

*Biased segmentation between large and small objects*: As the spatial resolution becomes coarse, geo-objects with diverse sizes (small and large) are segmented differently. As illustrated in **Table 2** and **3**, the large geo-objects (e.g. water, woodland) suffer slightly in performance, whereas the small objects (e.g. building) are influenced significantly.

*Coarse object boundary*: In urban scenarios, geo-objects located adjacently often present visually similar characteristics, which could lead to inaccurate segmentation results of object boundary. Such negative effect is becoming severe with detailed information loss in coarse resolution images. As illustrated in **Figure 7** and **9**, the boundary of buildings is easily confused with adjacent buildings at 0.25× spatial resolution.

### 4.2. Discussion of scale-aware feature representation

The comprehensive experiments demonstrate the superiority of our scale-aware feature representation for semantic segmentation of MSR remotely sensed images. Three vital factors ensure the competitive accuracies of our method. First, the proposed SFRM module models the invariant global spatial relationships to alleviate the contradiction between intact semantic content extraction and detailed information loss at coarse spatial resolutions. Commonly, a single-branched structure is used to capture the global spatial relationships of the networks. However, such an approach demonstrates weak adaptability to MSR images due to the fixed size of the branch feature. By contrast, our SFRM employs a dual-branched structure, where the global spatial relationships are modelled at different scales to adapt to the multi-resolution patterns. Second, the proposed DCFFM resolves the imbalanced segmentation quality of large and small objects. Traditional FPN fuses the high-level semantic features and low-level detailed features to capture multi-scale contexts. Although this can reduce the negative impact brought by the scale variation of geo-objects, the limited receptive field of extracted features restricts its representation capability severely in MSR images. Nevertheless, our DCFFM creates three large field connections to enrich the receptive field of semantic features, providing high-quality multi-scale contextual information. Moreover, our DCFFM employs weighted operations to aggregate multi-layer and multi-view features instead of fusing features directly, ensuring the generalization of the fused features. Third, building on the advantages of SFRM and DCFFM, our SaNet can capture the scale-aware feature for robust semantic segmentation of MSR images with the highest accuracy compared with all benchmark approaches.

*4.3. Application scenarios and model efficiency*

The main application scenario of the proposed method is multi-scale geo-object segmentation in MSR remotely sensed images, which can be applied to satellite sensors, aerial images and UAV drones captured at multiple scales. The reasons are: (1) With the advancement of sensor technology, remote sensing images are acquired at multiple resolutions at every point of the Earth. (2) Geo-objects within multi-resolution images are presented with a large variation in size and geometry. By learning scale-aware feature representation, our SaNet could pay equal attention to multi-scale objects, thereby segmenting geo-objects with complete and fine boundaries, demonstrating high accuracy and utility in such application scenarios. However, the computational cost from the combination of DCFFM and SFRM is increased in SaNet, with inevitably reduced computational efficiency. Our future research will, therefore, be devoted to designing an efficient and lightweight deep network to extract scale-aware features from MSR remotely sensed imagery.

## 5. Conclusions

Multi-resolution semantic segmentation is a challenging task due to the large variation in different objects and the information loss of fine details in MSR images. In this research, we present a scale-aware neural network (SaNet) for robust segmentation of MSR remotely sensed images using two novel modules, including a spatial feature recalibration module (SFRM) and a densely connected feature fusion module (DCFFM). Ablation studies indicate that both multi-scale representation and spatial relationship enhancement could increase the adaptability of the network to MSR images. The proposed SFRM module demonstrates superiority in characterising spatial relationships of the network compared to the SRM module, whereas the proposed DCFFM module captured high-quality multi-scale semantic information by merging various features. The combination of DCFFM and SFRM increased classification accuracy by learning scale-aware feature representation. Extensive experiments on three multi-resolution datasets (Landcover.ai, MSR Vaihingen and MSR Potsdam) demonstrates the strong cross-resolution generalisation capability of our SaNet compared with state-of-the-art benchmark approaches. Moreover, the proposed SFRM and DCFFM can be easily deployed and transplanted into any FCN-based segmentation network for precise segmentation of multi-resolution images automatically.

**Author Contributions:** This work was conducted in collaboration with all authors. Xiaoliang Meng defined the research theme and supervised the research work and provided experimental facilities. Libo Wang designed the semantic segmentation model and conducted the experiments. This manuscript was written by Libo Wang. Ce Zhang, Rui Li and Peter M. Atkinson revised the manuscript. Chenxi Duan checked the experimental results. All authors have read and agreed to the published version of the manuscript.

**Funding:** This research was funded by the National Natural Science Foundation of China (NSFC) under grant number 41971352, National Key Research and Development Program of China under grant number 2018YFB0505003.

**Acknowledgements**: The authors are very grateful to the many people who helped to comment on the article, and the Large Scale Environment Remote Sensing Platform (Facility No. 16000009, 16000011, 16000012) provided by Wuhan University, and the supports provided by Surveying and Mapping Institute Lands and Resource Department of Guangdong Province, Guangzhou. Special thanks to editors and reviewers for providing valuable insight into this article.

**Conflicts of Interest:** The authors declare no conflict of interest.

**Abbreviations:**

The following abbreviations are used in this manuscript:

| | |
|---|---|
| FSR | Fine Spatial Resolution |
| MSR | Multiple Spatial Resolutions |
| DCNNs | Deep Convolutional Neural Networks |
| MFF | Multi-level Feature Fusion |
| SPP | Spatial Pyramid Pooling |
| FCN | Fully Convolutional Neural Network |
| SVM | Support Vector Machine |
| RF | Random Forest |
| SIFT | Scale-invariant Feature Transformer |
| CRF | Conditional Random Field |
| SFRM | Spatial Feature Recalibration Module |
| DCFFM | Densely Connected Feature Fusion Module |
| FPN | Feature Pyramid Network |
| SRM | Spatial Relationship Module |
| SaNet | Scale-aware Neural Network |